\documentclass[11pt]{article}

\usepackage{acl}
\usepackage{amsfonts}
\usepackage{tcolorbox}
\usepackage{booktabs}
\usepackage{xcolor}
\usepackage{colortbl}
\definecolor{teacherblue}{RGB}{230, 240, 255}
\definecolor{studentpurple}{RGB}{240, 230, 255}
\definecolor{successgreen}{RGB}{230, 250, 230}
\definecolor{failred}{RGB}{255, 235, 235}
\usepackage{booktabs} 
\usepackage{multirow}
\usepackage{geometry}
\usepackage{cuted}
\usepackage{amsthm}
\usepackage{booktabs}
\usepackage{amssymb}
\usepackage{amsmath}
\usepackage{times}
\usepackage{latexsym}
\usepackage{graphicx}
\usepackage{subcaption}
\usepackage[T1]{fontenc}

\usepackage[utf8]{inputenc}
\usepackage{caption}
\usepackage{microtype}

\usepackage{inconsolata}

\usepackage{graphicx}

\author{
    \textbf{Dayu Wang}$^{1,2}$ \quad
    \textbf{Jiaye Yang}$^{1}$ \quad
    \textbf{Weikang Li}$^{1}$\thanks{Project leads and corresponding authors.} \quad
    \textbf{Jiahui Liang}$^{1}$ \quad
    \textbf{Yang Li}$^{1}$ \quad
    \textbf{Deguo Xia}$^{1}$ \quad
    \textbf{Jizhou Huang}$^{1}$\footnotemark[1] \\
    $^{1}$Baidu Inc. \qquad
    $^{2}$Nanyang Technological University \\
    \texttt{dayu001@e.ntu.edu.sg} \quad
    \texttt{yamseyoung@gmail.com} \\
    \texttt{wavejkd@pku.edu.cn} \quad
    \texttt{\{liangjiahui03, liyang164, xiadeguo, huangjizhou01\}@baidu.com}
}

\begin{document}

\title{Student Guides Teacher: Weak-to-Strong Inference via Spectral Orthogonal Exploration}
\maketitle

\begin{abstract}
Large Language Models (LLMs) often suffer from ``Reasoning Collapse'' on challenging mathematical reasoning tasks, where stochastic sampling produces lexical variations of the same erroneous logic rather than genuine semantic exploration. We observe that failed reasoning traces are often associated with a low-rank bias manifold in the model's hidden-state geometry, which reduces exploration toward corrective solution directions. To address this, we propose \textbf{Spectral Orthogonal Exploration (SOE)}, a geometric inference framework under a ``Student Guides Teacher'' paradigm. Instead of using a weak auxiliary agent for imitation, SOE uses it as an \textit{orthogonal probe} to introduce semantically heterogeneous reasoning signals into the teacher's orthogonal complement of its dominant subspace. This intervention steers the teacher toward more diverse reasoning trajectories and improves exploration beyond standard sampling. Experiments on mathematical benchmarks show that SOE improves average accuracy by \textbf{62.4\%} and average sampling efficiency by \textbf{113.7\%} over baseline methods, suggesting that geometric interventions can be effective for mitigating reasoning collapse in mathematical reasoning. We further provide preliminary evidence that SOE is also effective on logic and code generation benchmarks. Code is available at \url{https://github.com/dayuwang401/spectral-orthogonal-exploration}.
\end{abstract}

\begin{figure}[!t] 
    \centering
    \includegraphics[width=0.8\linewidth]{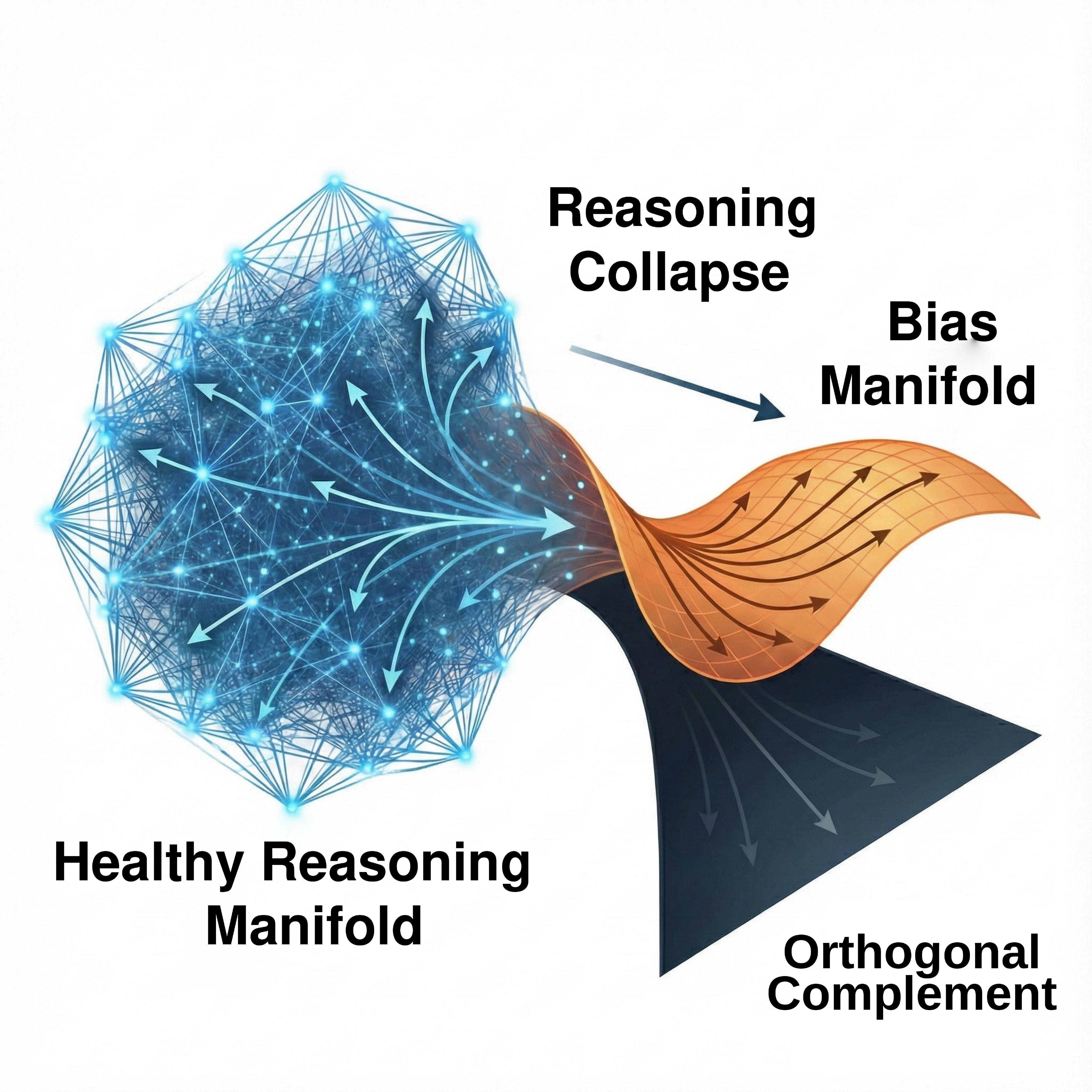}
\caption{Geometric interpretation of \textbf{Reasoning Collapse}. We illustrate reasoning collapse as a transition of the state space from a high-dimensional \textbf{Healthy Reasoning Manifold} to a low-rank \textbf{Bias Manifold}. This concentration may reduce exploration toward corrective solutions in the \textbf{Orthogonal Complement}, as the trajectory becomes confined to a lower-dimensional region.}
\label{fig:reasoning_collapse}    
\end{figure}

\section{Introduction}
\begin{figure*}[t]
    \centering
    
    \begin{subfigure}{\textwidth}
        \centering
        \includegraphics[width=0.90\linewidth]{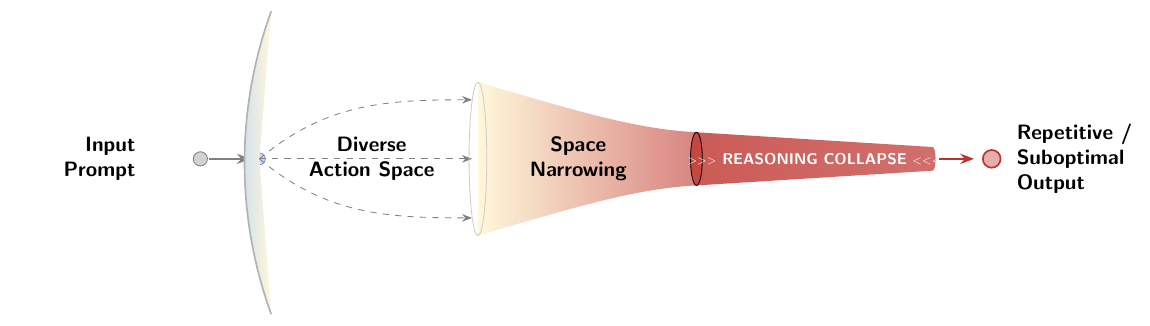} 
        \caption{Reasoning Collapse in Vanilla Inference.}
        \label{fig:collapse}
    \end{subfigure}

    \vspace{0.5cm} 

    \begin{subfigure}{\textwidth}
        \centering
        \includegraphics[width=0.90\linewidth]{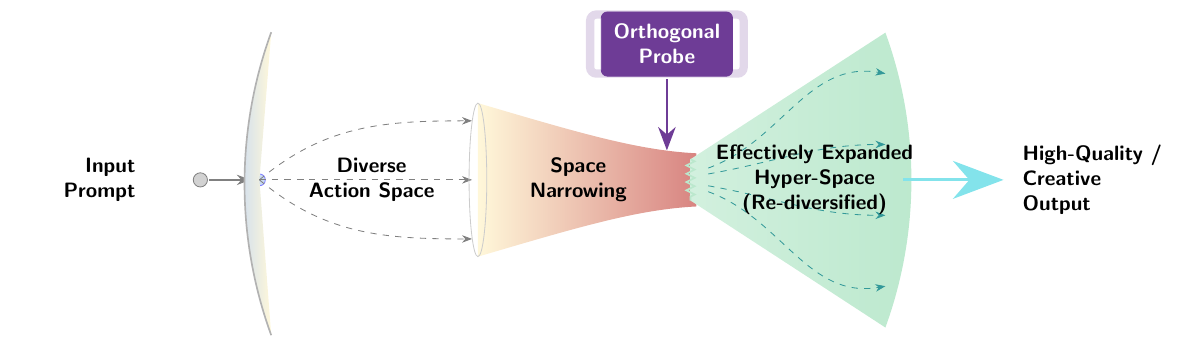} 
        \caption{Ours: Spectral Orthogonal Exploration (SOE) Intervention.}
        \label{fig:soe_intervention}
    \end{subfigure}

    \caption{Mechanism of Spectral Orthogonal Exploration (SOE). (a) Standard inference leads to space narrowing and reasoning collapse. (b) To counteract this, we introduce an \textbf{Orthogonal Probe} as a geometric intervention. This force effectively disrupts the low-rank confinement and diversifies the reasoning trajectory, expanding the hyper-space to access high-quality solutions.}
    \label{fig:motivation}
\end{figure*}

    

Current Large Language Models (LLMs) exhibit a paradox in complex reasoning: while they demonstrate emergent capabilities driven by scaling laws~\cite{kaplan2020scaling,wei2022emergentabilitieslargelanguage,hoffmann2022trainingcomputeoptimallargelanguage,deepseekai2025deepseekv3technicalreport,deepseekai2025deepseekr1incentivizingreasoningcapability,deepseekai2025deepseekv32pushingfrontieropen,bubeck2025earlyscienceaccelerationexperiments}, they frequently encounter a recurring failure mode known as \textit{Reasoning Collapse}~\cite{shojaee2025illusionthinkingunderstandingstrengths,zhao2025chainofthoughtreasoningllmsmirage,guo2025mathematicalprooflitmustest,yamin2025failuremodesllmscausal}. When facing challenging mathematical reasoning, logic, or code generation tasks, model performance does not decay gracefully but often stagnates abruptly~\cite{petrov2025proofbluffevaluatingllms}. We hypothesize that this phenomenon is associated with a \textbf{geometric collapse of the representation space}~\cite{li2025tracingrepresentationgeometrylanguage,laurent2023featurecollapse,fan2025combattingdimensionalcollapsellm}, as illustrated in Figure~\ref{fig:reasoning_collapse}.

Specifically, we posit the \textit{Low-Rank Manifold Hypothesis}: as a strong model becomes overly confident in an erroneous reasoning path, its internal hidden states may become concentrated in a low-dimensional \textbf{Bias Manifold}~\cite{minegishi2025topologyreasoningunderstandinglarge,bazarova2025hallucinationdetectionllmstopological,phillips2025geometricuncertaintydetectingcorrecting,park2025steerllmlatentshallucination}. In this collapsed state, the model's covariance matrix becomes increasingly ill-conditioned, retaining high variance primarily in directions aligned with its current bias, while variance in corrective directions is substantially reduced~\cite{dohmatob2024modelcollapsedemystifiedcase}.

To address this geometric limitation, we introduce the concept of \textbf{Spectral Orthogonal Exploration (SOE)}, illustrated in Figure~\ref{fig:soe_intervention}. Our approach is grounded in a fundamental linear algebraic principle: escaping a restricted subspace requires a vector with a non-zero component in its orthogonal complement. We propose a counter-intuitive \textbf{Student Guides Teacher} paradigm, where a weaker Student model acts not as a supervisor, but as an \textbf{Orthogonal Probe}. Crucially, the Student's utility derives not from its predictive accuracy, but from its structural heterogeneity. Since the Student has not converged toward the same Bias Manifold as the Teacher, its reasoning trajectories---though potentially suboptimal---may retain significant components in directions underexplored by the Teacher. By calculating the \textbf{Orthogonal Projection Residual} of the Student's outputs relative to the Teacher's dominant eigenspace, we obtain an intervention signal that steers the Teacher away from its current biased trajectory.

\section{Related Work}
Traditional alignment and exploration methods, such as high-temperature sampling~\cite{holtzman2020curiouscaseneuraltext} or diverse prompting, often struggle to rectify the structural pathology of reasoning collapse~\cite{zhang2025siren,lu2023survey,gudibande2023false,casper2023open,ethayarajh2019contextual,shumailov2023curse}. Because these methods operate primarily in the probabilistic token space, they implicitly assume that lexical diversity equates to semantic exploration~\cite{shi2025semanticguideddiversedecodinglarge,shypula2025evaluatingdiversityqualityllm}.

\begin{figure*}[t] 
    \centering
    \begin{subfigure}{0.24\textwidth}
        \centering
        \includegraphics[width=\linewidth]{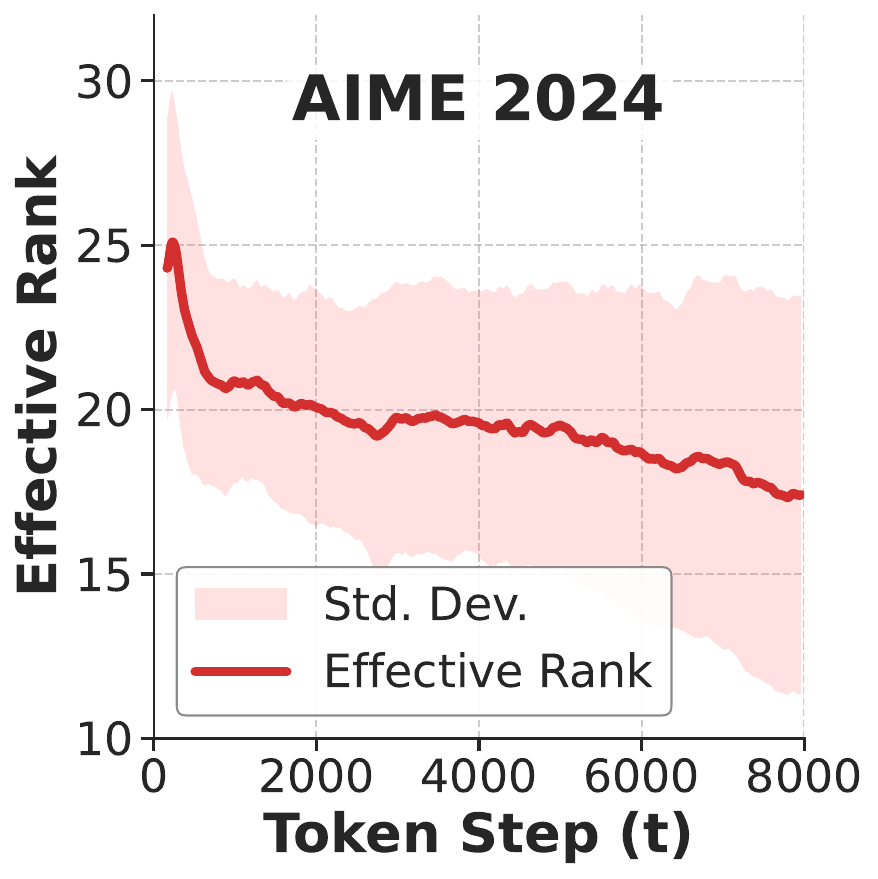}
        \label{fig:collapse_aime24}
    \end{subfigure}
    \hfill 
    \begin{subfigure}{0.24\textwidth}
        \centering
        \includegraphics[width=\linewidth]{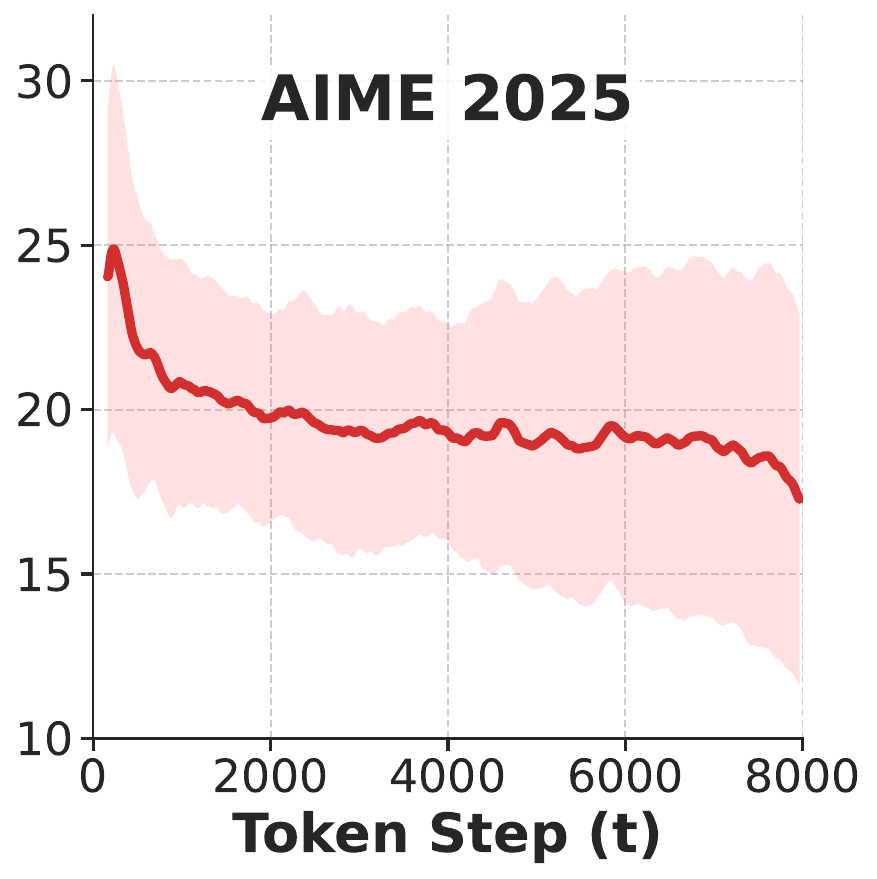}
        \label{fig:collapse_aime25}
    \end{subfigure}
    \hfill
    \begin{subfigure}{0.24\textwidth}
        \centering
        \includegraphics[width=\linewidth]{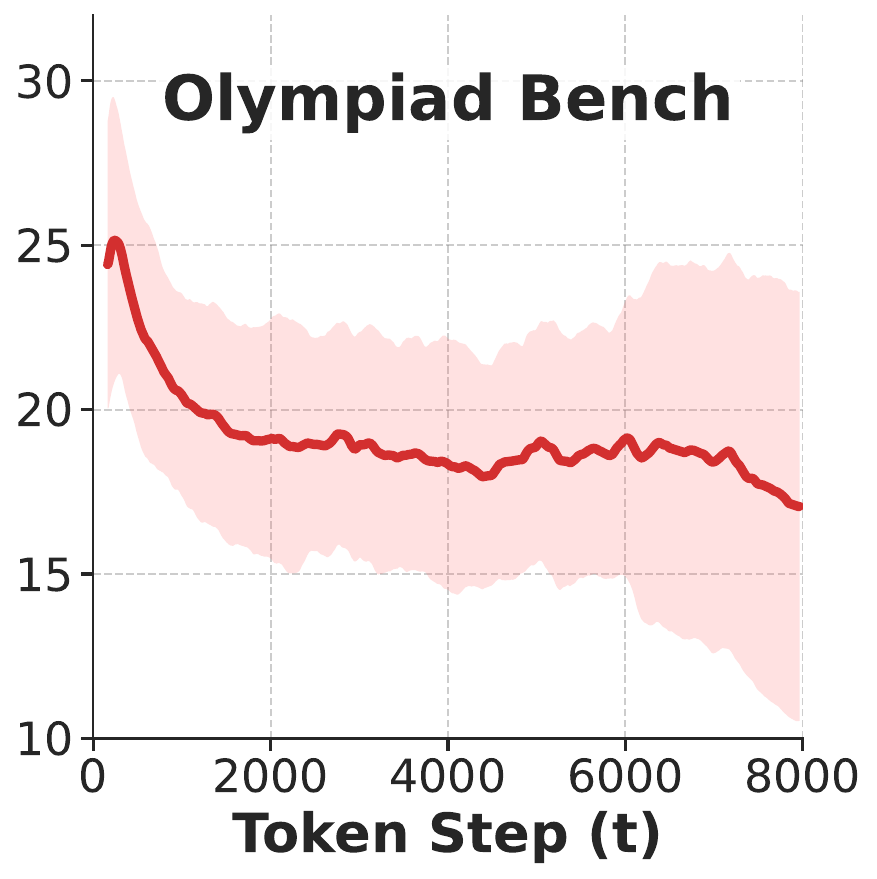}
        \label{fig:collapse_oly}
    \end{subfigure}
    \hfill
    \begin{subfigure}{0.24\textwidth}
        \centering
        \includegraphics[width=\linewidth]{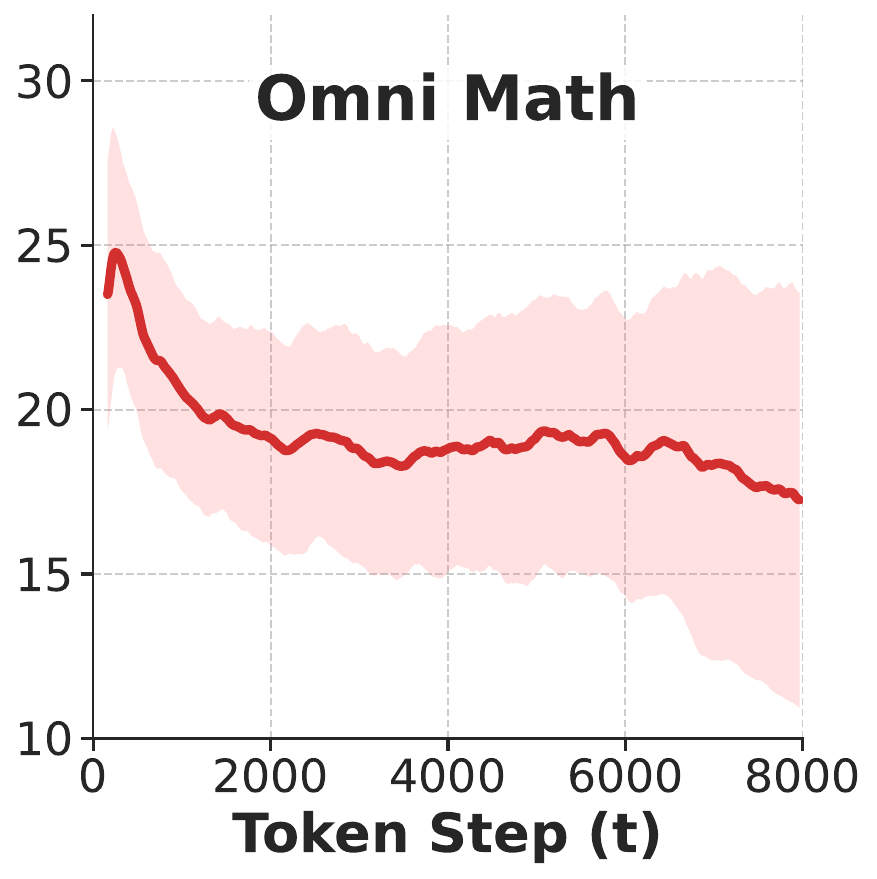}
        \label{fig:collapse_omni}
    \end{subfigure}
    
    \vspace{-2mm} 
    \caption{\textbf{Universality of Reasoning Collapse across Datasets.} 
    The dynamics of Effective Rank during the reasoning process on AIME 24, AIME 25, Olympiad Bench and Omni-Math. 
    Across all benchmarks, we observe a consistent spectral degeneration pattern: as the token steps increase, the effective rank of the covariance matrix decays, consistent with increasing concentration into a low-dimensional bias manifold across benchmarks.}
    \label{fig:universal_collapse}
    \vspace{-3mm} 
\end{figure*}

However, from a geometric perspective, stochastic sampling on a collapsed manifold is similar to \textbf{Brownian motion restricted to a 2D plane}: no matter how much noise is injected, the trajectory cannot escape to the third dimension~\cite{seddik2024badtrainingsyntheticdata,joseph2025lambdaskipconnectionsarchitecturalcomponent,farghly2025diffusionmodelsmanifoldhypothesis}. Consequently, the model generates fluent and diverse yet logically redundant fallacies, remaining effectively trapped within a spectral "blind spot" that stochasticity alone cannot illuminate.

\section{Motivation: The Geometry of Reasoning Collapse}

To understand the nature of reasoning failure, we move beyond surface-level token statistics to analyze the geometric properties of the model's representation space. While reasoning collapse is often identified by repetitive text or logical loops, we hypothesize that these patterns may reflect an underlying geometric phenomenon, which we refer to as \textbf{Spectral Collapse}.

\subsection{The Low-Rank Manifold Hypothesis}
Consider the inference process of a Large Language Model as a discrete dynamical system. We define the hidden state trajectory of a reasoning chain up to step $t$ as $H_t = [h_1, h_2, \dots, h_t] \in \mathbb{R}^{t \times d}$, formed by stacking the hidden state vectors. Specifically, the $i$-th row represents the hidden state $h_i \in \mathbb{R}^{1 \times d}$ at step $i$.
As the model commits to a specific reasoning path, its internal representations may become increasingly concentrated within a lower-dimensional subspace. To describe the dimension of the subspace, we analyze the \textit{Local Covariance Matrix} $\Sigma_t$ of the recent window of $k$ tokens:
\begin{equation}
    \Sigma_t = \frac{1}{k-1} \sum_{i=t-k+1}^{t} (h_i - \mu_t)^\top(h_i - \mu_t)
\end{equation}
where $\mu_t$ is the local mean.
When the effective dimensionality of this subspace is low, nearby hidden states become more similar, and the \textit{Local Covariance Matrix} correspondingly exhibits lower effective rank. Based on this, we posit the \textbf{Low-Rank Manifold Hypothesis}: during reasoning collapse, the effective rank of $\Sigma_t$ tends to decrease, indicating that the reasoning process becomes increasingly concentrated in a low-dimensional \textbf{Bias Manifold} $\mathcal{M}$ \cite{gao2019representation}.

\subsection{Quantifying Spectral Collapse}
Based on the analysis above, we use the Effective Rank (\textit{EffRank}) \cite{roy2007effective} to quantify the dimensionality of this bias manifold:
\begin{equation}
    \textit{EffRank}(\Sigma_t) = \exp\left( -\sum_{j=1}^d \tilde{\sigma}_j \log \tilde{\sigma}_j \right)
\end{equation}
Here, $\tilde{\sigma}_j = \lambda_j / \sum_{i=1}^d \lambda_i$ represents the normalized singular values.
We interpret $\textit{EffRank}(\Sigma_t) \ll d$ as evidence of substantial geometric concentration in the representation space. In this state, variance in directions within the orthogonal complement of the dominant subspace is substantially reduced. Consequently, even with stochastic sampling (temperature $T > 0$), generated trajectories may remain concentrated within the Bias Manifold, reducing exploration toward corrective solutions in the orthogonal complement.

\subsection{Empirical Analysis}

\begin{figure*}[t]
    \centering
    \includegraphics[width=\textwidth]{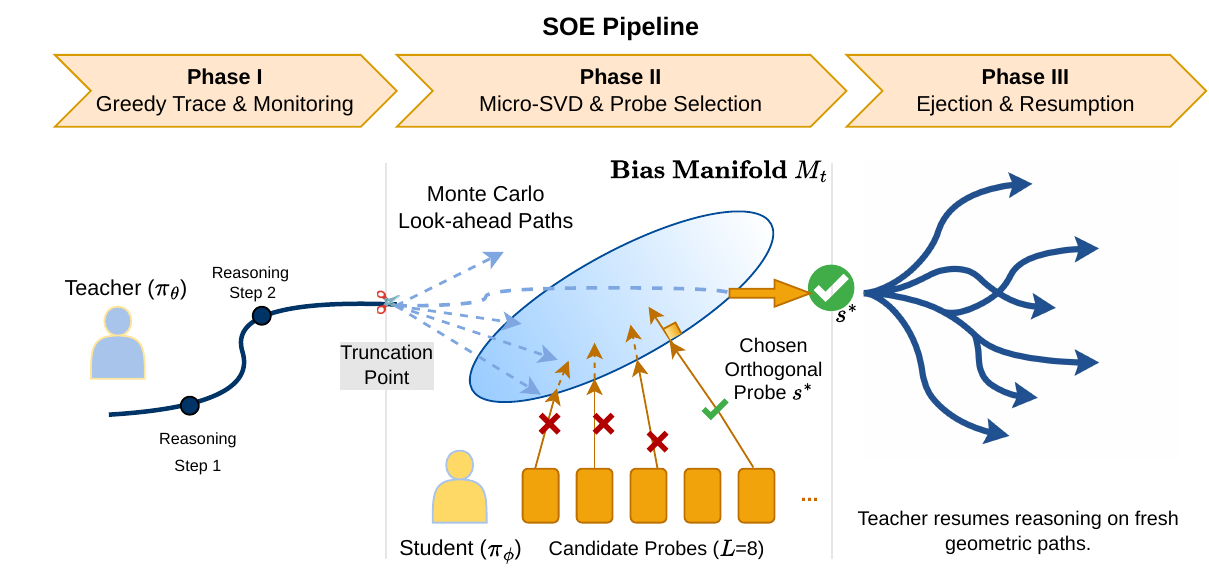}
    \caption{The pipeline of Spectral Orthogonal Exploration (SOE). 
A greedy reasoning trace from the Teacher is first monitored and truncated at candidate collapse points. 
At each truncation point, Monte Carlo samples are used to estimate the Teacher's local bias manifold via Micro-SVD, and the Student generates multiple short candidate probes. 
The probe with the largest orthogonal residual in the Teacher's latent space is selected and stitched into the context. 
The Teacher then resumes reasoning from the modified prefix, thereby exploring geometric directions outside its original bias manifold.}
    \label{fig:soe_framework}
\end{figure*}

To examine this hypothesis, we conducted a pilot study across four challenging mathematical benchmarks: AIME 24 \cite{patel2024aimeaioptimizationmultiple}, AIME 25 \cite{petrov2025proofbluffevaluatingllms}, Olympiad Bench \cite{he2024olympiadbench} and Omni-Math \cite{gao2024omnimathuniversalolympiadlevel}. We use \texttt{Qwen3-4B-Instruct-2507} \cite{qwen3technicalreport} to sample reasoning chains with a high temperature ($T=0.7$). To specifically isolate the phenomenon of reasoning collapse, we curate a subset of trajectories that are both factually incorrect and excessively verbose (exceeding 4,000 tokens). For each selected trajectory, we computed the Effective Rank of the covariance matrix of the final hidden states using a sliding window of size $k=64$.

As illustrated in Figure~\ref{fig:universal_collapse}, we observe a substantial decrease in effective rank over the course of long, erroneous reasoning chains. Despite a theoretical full rank of $d=2560$, the effective dimensionality consistently decays as generation progresses. This quantitative evidence is consistent with our hypothesis that reasoning loops are associated with concentration into a low-dimensional bias manifold, which may reduce the geometric degrees of freedom available for divergent exploration. These observations motivate the use of a geometric intervention to encourage broader exploration.

To further substantiate these findings, we provide an extended analysis in Appendix~\ref{sec:Reasoning Collapse Evidence}. There, we provide additional evidence that similar collapse patterns can be observed across multiple model sizes.

\section{Methodology: The SOE Framework}

Motivated by the observed geometric decay, we develop the \textbf{Spectral Orthogonal Exploration (SOE)} framework. This section describes the SOE algorithm for identifying reasoning collapse and intervening during the inference process.

\subsection{Overview of the Geometric Interaction}
The framework operates in a post-hoc intervention setting. As shown in Figure~\ref{fig:soe_framework}, the process consists of three stages:

\begin{enumerate}
    \item \textbf{Greedy Trace and Monitoring:} We first employ the Teacher $\pi_\theta$ to generate a complete reasoning chain via greedy decoding. We then collect trajectories that yield incorrect answers, treating these erroneous paths as candidate instances of collapsed reasoning. These trajectories are partitioned at critical reasoning milestones (e.g., the onset of logical steps). This procedure yields a set of prefix trajectories, allowing us to probe the model's latent state at various depths of the reasoning process.
    
    \item \textbf{Micro-SVD and Probe Selection:} At each truncation point $t_i$, we estimate the local bias manifold $\mathcal{M}_t$ using Monte Carlo look-ahead simulations and Micro-SVD to identify the dominant subspace associated with the Teacher's current reasoning trajectory. For each truncation point, the Student $\pi_\phi$ generates candidate probe sequences (fixed at $L=8$ tokens). We calculate the orthogonality of these candidates relative to the specific manifold $\mathcal{M}_t$ at that step, selecting the optimal probe $s^*_t$ that maximizes the orthogonality score.
    
    \item \textbf{Intervention and Resumption:} The previous process yields distinct intervened contexts (the truncated histories stitched with their respective orthogonal probes). The Teacher $\pi_\theta$ then resumes reasoning via sampling from each of these modified contexts, allowing it to continue exploration from three different geometric intervention points.
\end{enumerate}

A detailed example of the algorithm is provided in Appendix~\ref{app:soe_example}.

\subsection{Manifold Estimation}
\label{sec:micro_svd}

To estimate the bias manifold of the Teacher, we assume the Teacher's internal representations at truncation point $t$ follow a local distribution with covariance $\boldsymbol{\Sigma_t}$. Our goal is to obtain an \textit{unbiased estimate} of the principal components of $\boldsymbol{\Sigma_t}$.

\paragraph{Unbiased Covariance Estimator.}
Given the context $x_{<t}$ at the truncation point $t$, we employ a Monte Carlo method to sample $N$ stochastic look-ahead trajectories $\{ \tau_1, \dots, \tau_N \}$ from the Teacher $\pi_\theta(\cdot|x_{<t})$. 
Let $\mathbf{h}_i \in \mathbb{R}^d$ be the aggregated hidden state of the $i$-th trajectory. We calculate the sample mean $\hat{\boldsymbol{\mu}} = \frac{1}{N}\sum_{i=1}^N \mathbf{h}_i$ and construct the centered state matrix $\mathbf{H} \in \mathbb{R}^{d \times N}$:
\begin{equation}
    \mathbf{H} = \left[ \mathbf{h}_1 - \hat{\boldsymbol{\mu}}, \dots, \mathbf{h}_N - \hat{\boldsymbol{\mu}} \right]
\end{equation}
Theoretically, the sample covariance matrix $\hat{\mathbf{C}} = \frac{1}{N-1}\mathbf{H}\mathbf{H}^T$ serves as an unbiased estimator of the true population covariance $\boldsymbol{\Sigma_t}$. By analyzing the spectrum of $\hat{\mathbf{C}}$, we estimate the directions of maximum variance associated with the Teacher's current reasoning trajectory.

\paragraph{The Micro-SVD Algorithm.}
Directly computing eigenvectors for $\hat{\mathbf{C}} \in \mathbb{R}^{d \times d}$ is computationally prohibitive. We leverage the SVD duality by computing the Gram matrix $\mathbf{G} \in \mathbb{R}^{N \times N}$ of our Monte Carlo samples:
\begin{equation}
    \mathbf{G} = \mathbf{H}^T \mathbf{H}
\end{equation}
Solving the eigen-problem for the smaller matrix $\mathbf{G}\mathbf{v}_k = \lambda_k \mathbf{v}_k$ allows us to recover the principal components $\mathbf{u}_k$ of the Teacher's local subspace:
\begin{equation}
    \mathbf{u}_k = \frac{1}{\sqrt{\lambda_k}} \mathbf{H} \mathbf{v}_k
\end{equation}
We define the \textbf{Bias Manifold} $\mathcal{M}_t$ as the subspace spanned by the top-$k$ eigenvectors $\mathbf{U}_{\parallel} = \{ \mathbf{u}_1, \dots, \mathbf{u}_k \}$. Directions in the orthogonal complement of this subspace provide candidate directions for intervention and exploration.

\subsection{Probe Selection: Orthogonal Latent Stitching (OLS)}

With the manifold $\mathcal{M}_t$ identified, we employ Orthogonal Latent Stitching to intervene on the model's current reasoning trajectory.

\paragraph{Heterogeneous Probing.}
The Student model $\pi_\phi$ generates a set of candidate sequences $\mathcal{C}_{Student} = \{ s_1, \dots, s_M \}$, each with a fixed length of $L=8$ tokens. 
These short sequences act as geometric probe vectors. To evaluate their geometry, we map them into the Teacher's latent space via a forward pass: $\mathbf{z}_j = \textit{Forward}_{\pi_\theta}(s_j \mid x_{<t})$.

\paragraph{Residual Maximization.}
We compute the projection of each candidate $\mathbf{z}_j$ onto the Teacher's bias manifold using the projection operator $\mathbf{P}_{\parallel} = \mathbf{U}_{\parallel}\mathbf{U}_{\parallel}^T$. The orthogonal residual $\mathbf{r}_j$ represents the component of the Student's probe that lies in the orthogonal complement of the Teacher's dominant subspace:
\begin{equation}
    \mathbf{r}_j = ( \mathbf{I} - \mathbf{P}_{\parallel} ) (\mathbf{z}_j - \hat{\boldsymbol{\mu}})
\end{equation}
We select the candidate $s^*$ that maximizes the Orthogonality Score, defined as the normalized energy of this residual, where $\epsilon > 0$ is a small constant introduced to avoid division by zero:
\begin{equation}
    s^* = \arg\max_{s_j \in \mathcal{C}_{Student}} \frac{\| \mathbf{r}_j \|_2}{\| \mathbf{z}_j - \hat{\boldsymbol{\mu}} \|_2 + \epsilon}
\end{equation}
This orthogonal token sequence $s^*$ is then stitched to the truncated context, providing an intervention that steers the Teacher toward exploration from a new geometric direction outside its original bias manifold.

\section{Experiments}
\label{sec:experiments}

To validate the theoretical claims of Spectral Orthogonal Exploration (SOE), we conducted extensive experiments focused on high-difficulty mathematical reasoning tasks. 

\subsection{Experimental Setup}
\begin{figure*}[t]
    \centering
    
    
    
    \resizebox{\textwidth}{!}{%
        \begin{tabular}{lcccccc}
        \toprule
        \textbf{Method} & \textbf{AIME 24} & \textbf{AIME 25} & \textbf{MATH-500} & \textbf{Olympiad Bench} & \textbf{Omni-Math (Hard)} & \textbf{Average} \\
        \midrule
        Baseline (Self-Consistency) & 38.5\% & 35.3\% & 33.7\% & 11.7\% & 14.5\% & 26.7\% \\
        \textbf{Ours (SOE)} & \textbf{76.9\%} & \textbf{70.6\%} & \textbf{45.9\%} & \textbf{15.5\%} & \textbf{20.8\%} & \textbf{45.9\%} \\
        \midrule
        Relative Improvement & +99.7\% & +100.0\% & +36.2\% & +32.5\% & +43.4\% & +62.4\% \\
        \bottomrule
        \end{tabular}
    }
    \captionsetup{width=\textwidth, justification=justified} 
    \captionof{table}{Pass@16 accuracy on the \textbf{Difficult Subset} (problems where Teacher's greedy decoding failed). SOE demonstrates a consistent improvement over the Baseline, confirming that directed orthogonal exploration is superior to standard self-consistency sampling. All the results are sampled under identical conditions, with the same prompt, $temperature = 0.7$ and a maximum context (sampling) length capped at 8192 tokens.}
    \label{tab:performance_comparison}

    \vspace{1.5em} 

    \includegraphics[width=\textwidth]{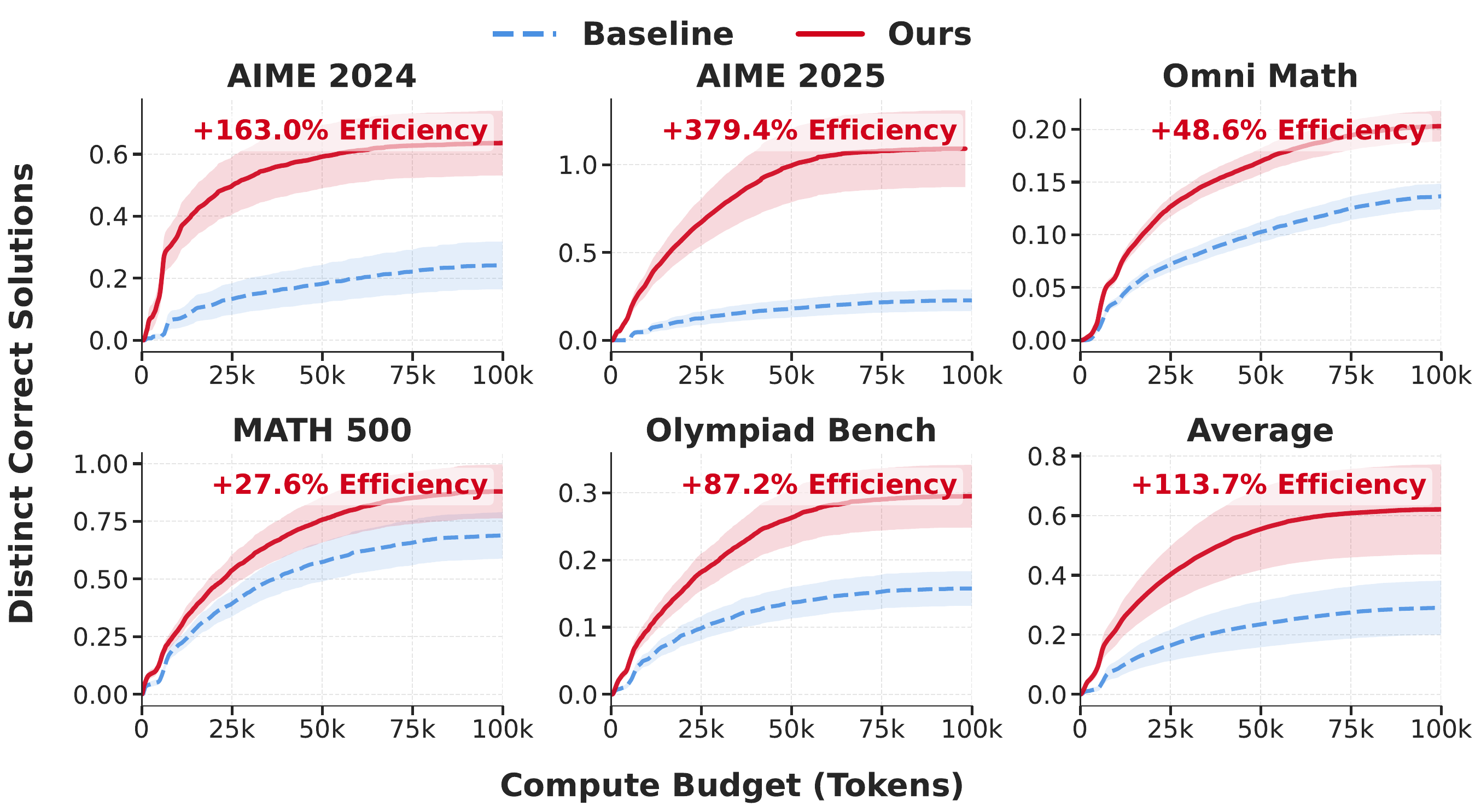}
    \captionof{figure}{Semantic Exploration Efficiency. The curves depict the cumulative number of distinct correct solutions discovered as a function of the compute budget (total tokens). SOE significantly outperforms the Baseline by maintaining a near-linear discovery rate, whereas the Baseline exhibits logarithmic saturation.}
    \label{fig:placeholder}

\end{figure*}

\paragraph{Models.} We adopt a ``Weak-Student and Strong-Teacher'' configuration to validate our proposal:
We use \texttt{Qwen3-4B-Instruct-2507} as the Teacher, and use \texttt{Gemma-3-4B-IT} \cite{gemma_2025} as the Student. Crucially, the Student's different architecture and training data ensure that its bias manifold is not perfectly aligned with the Teacher's.
\begin{table}[h]
    \centering
    
    \label{tab:model_comparison}
    \begin{tabular}{lcc}
        \toprule
        \textbf{Dataset} & \textbf{Teacher} & \textbf{Student} \\
        \midrule
        AIME 2024       & \textbf{50.0\%} & 13.3\% \\
        AIME 2025       & \textbf{33.7\%} & 6.7\% \\
        MATH-500        & \textbf{86.8\%} & 71.2\% \\
        Olympiad Bench   & \textbf{45.3\%} & 27.7\% \\
        Omni-Math(Hard)       & \textbf{18.3\%} & 4.2\% \\
        \bottomrule
    \end{tabular}
    \caption{Comparison of fundamental capabilities between Teacher and Student models on mathematical benchmarks. Both the Teacher and Student are evaluated under identical conditions, with the same prompt, greedy decoding and a maximum context length capped at 8192 tokens.}
\end{table}

\paragraph{Datasets.} We evaluate performance on five challenging mathematical benchmarks: AIME 2024, AIME 2025, MATH-500 \cite{hendrycks2021measuring}, Olympiad Bench and Omni-Math. Given the strong capabilities of the Teacher model, we specifically selected problems from Omni-Math with a difficulty level greater than or equal to 8.

\paragraph{Evaluation.}
For all benchmarks, model outputs are post-processed using a combination of regular-expression-based normalization and the \texttt{MathEvaluator} library to extract and verify final answers against ground truth.

\begin{table*}[t]
    \centering
    \small
    \begin{tabular}{llccc}
        \toprule
        \textbf{Teacher} & \textbf{Student} & \textbf{Benchmark} & \textbf{Baseline} & \textbf{Ours (SOE)} \\
        \midrule
        Qwen3-4B-Instruct-2507 & DeepSeek-R1-Distill-Qwen-7B & AIME 2024 & 38.46\% & \textbf{76.92\%} \\
        Qwen3-4B-Instruct-2507 & DeepSeek-R1-Distill-Qwen-7B & AIME 2025 & 35.29\% & \textbf{64.70\%} \\
        Qwen3-4B-Instruct-2507 & Mistral-7B-Instruct-v0.3    & AIME 2024 & 38.46\% & \textbf{69.23\%} \\
        Qwen3-4B-Instruct-2507 & Mistral-7B-Instruct-v0.3    & AIME 2025 & 35.29\% & \textbf{58.82\%} \\
        Qwen3-8B               & Gemma-3-4B-IT               & AIME 2024 & 14.29\% & \textbf{42.86\%} \\
        Qwen3-8B               & Gemma-3-4B-IT               & AIME 2025 & 4.35\%  & \textbf{26.09\%} \\
        Qwen3-32B              & Gemma-3-4B-IT               & AIME 2024 & 11.11\% & \textbf{55.56\%} \\
        Qwen3-32B              & Gemma-3-4B-IT               & AIME 2025 & 21.74\% & \textbf{34.78\%} \\
        \bottomrule
    \end{tabular}
    \caption{Cross-family and scale generalization of SOE. The method remains effective with heterogeneous Teacher--Student pairs and larger Teacher models.}
    \label{tab:generalization}
\end{table*}

\paragraph{Baselines and Implementation Details.} We compare two experimental settings:
\begin{itemize}
    \item \textbf{Baseline (Self-Consistency):} Standard sampling with the Teacher model ($T=0.7$). This represents the control group for stochastic exploration within the Bias Manifold.
    \item \textbf{Ours (SOE):} We generate $N=8$ candidates from the Student ($T=1.0$), compute their embeddings using the Teacher model, and mechanically stitch the candidate maximizing the Orthogonality Score ($\Omega(s) \to 1$) into the Teacher's context. The subsequent reasoning chain is completed by the Teacher ($T=0.7$).
\end{itemize}

\subsection{Main Results}

We evaluate the solution accuracy using \textbf{Pass@$16$}. As shown in Table~\ref{tab:performance_comparison}, SOE significantly outperforms the Self-Consistency baseline across all datasets on the difficult subset.

We further compare SOE against a strong step-level Best-of-$N$ baseline using a Process Reward Model (PRM). As shown in Table~\ref{tab:prm_baseline}, SOE consistently surpasses the PRM baseline on AIME 2024, AIME 2025, and MATH-500 under matched sampling settings, indicating that the gains are not solely attributable to stronger step selection.
\begin{table}[h]
    \centering
    \small
    \resizebox{\columnwidth}{!}{%
    \begin{tabular}{lccc}
        \toprule
        \textbf{Dataset} & \textbf{PRM Best-of-$N$} & \textbf{Ours (SOE)} & \textbf{Rel. Imp.} \\
        \midrule
        AIME 2024 & 69.23\% & \textbf{76.90\%} & +11.08\% \\
        AIME 2025 & 58.82\% & \textbf{70.60\%} & +20.03\% \\
        MATH-500  & 40.98\% & \textbf{45.90\%} & +12.01\% \\
        \bottomrule
    \end{tabular}%
    }
    \caption{Comparison against a strong step-level Best-of-$N$ baseline using a Process Reward Model (PRM). All sampling positions and the number of subsequent trajectories are matched to SOE.}
    \label{tab:prm_baseline}
\end{table}

To evaluate generalizability beyond a single 4B Teacher--Student pair, we further test heterogeneous cross-family combinations and larger Teacher models. Table~\ref{tab:generalization} shows that SOE remains effective when using DeepSeek-R1-Distill-Qwen-7B and Mistral-7B-Instruct-v0.3 \cite{jiang2023mistral7b} as Student models, and when scaling the Teacher from Qwen3-4B to Qwen3-8B and Qwen3-32B.

\subsection{Control Experiments on Rank Decay}

To disentangle rank decay from mere verbosity or looping, we conduct matched-control experiments on AIME 2025 with 16 samples per problem. Specifically, we compare the percentage drop in Effective Rank between the first 500 tokens and the final 500 tokens across four cohorts: short/long and correct/wrong traces.

\begin{table}[h]
    \centering
    \small
    \begin{tabular}{lc}
        \toprule
        \textbf{Trace Category} & \textbf{Avg. Rank Drop} \\
        \midrule
        Short \& Correct ($<4$k tokens) & 4.82\% \\
        Short \& Wrong ($<4$k tokens)   & 19.65\% \\
        Long \& Correct ($>5$k tokens)  & 13.14\% \\
        Long \& Wrong ($>5$k tokens)    & 27.04\% \\
        \bottomrule
    \end{tabular}
    \caption{Matched-control analysis on AIME 2025. Effective-rank drop is computed by comparing the first 500 tokens and the final 500 tokens of each trajectory.}
    \label{tab:matched_rank}
\end{table}

As shown in Table~\ref{tab:matched_rank}, long incorrect traces suffer more than double the rank decay of long correct traces, while short incorrect traces already exhibit a substantial drop even without extended repetition. These results suggest that rank decay is not merely a byproduct of verbosity, but is closely associated with reasoning failure.

\subsection{Exploration Efficiency}
\label{sec:exploration_efficiency}

\begin{figure*}[t]
    \centering
    \includegraphics[width=\textwidth]{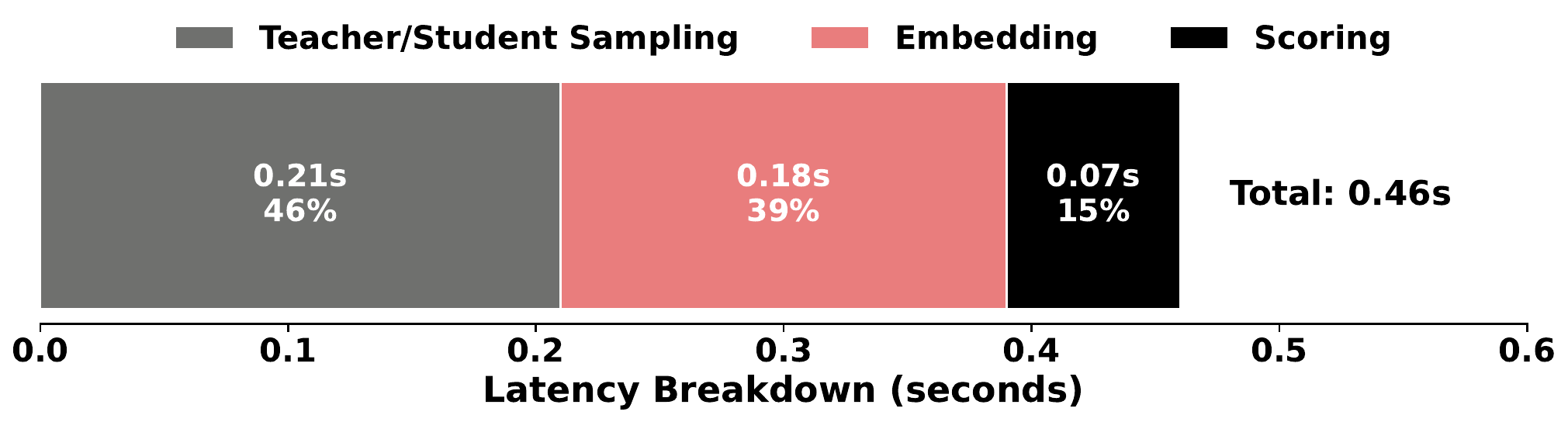}
    \caption{Latency breakdown of a single SOE intervention step evaluated on the AIME 2025 benchmark. The process was measured on an NVIDIA A800 GPU using the vLLM pipeline.}
    \label{Time}
\end{figure*}

Given that Spectral Orthogonal Exploration (SOE) introduces additional inference cost, it is important to assess whether this overhead translates into more effective exploration. To this end, we compare SOE with the Self-Consistency baseline in terms of \textbf{Semantic Exploration Efficiency}, which measures how efficiently a method discovers \textit{semantically distinct} correct reasoning trajectories under a fixed generation budget. This metric is particularly relevant for \textbf{data synthesis}, where semantically diverse correct traces are more useful than redundant paraphrases of the same solution.

\paragraph{Metric Definition.}
We use \texttt{jina} embeddings \cite{günther2023jina} to represent generated correct solution traces. A correct solution is counted as ``new'' only if its cosine similarity to all previously discovered correct solutions is below a threshold of $\tau=0.95$, thereby filtering out trivial paraphrases.

As illustrated in Figure~\ref{fig:placeholder}, SOE consistently achieves higher exploration efficiency than the baseline.

\begin{table}[t]
    \centering
    \small
    \setlength{\tabcolsep}{3pt}
    \begin{tabular}{lccc}
        \toprule
        \textbf{Benchmark} & \textbf{Baseline} & \textbf{Ours} & \textbf{Ours (Time-Norm.)} \\
        \midrule
        AIME 2024  & 42.86\% & 77.14\% & 68.39\% \\
        AIME 2025  & 36.00\% & 72.00\% & 63.83\% \\
        MATH-500   & 35.12\% & 44.64\% & 39.57\% \\
        Olympiad Bench   & 9.49\%  & 14.09\% & 12.49\% \\
        Omni-Math  & 14.39\% & 20.45\% & 18.13\% \\
        \bottomrule
    \end{tabular}
    \caption{Sample accuracy under wall-clock normalization. Following the measured 12.8\% runtime overhead of SOE under vLLM, we divide the SOE count by $1.128$ for a strict time-normalized comparison.}
    \label{tab:time_normalized}
\end{table}

As detailed in Figure~\ref{Time}, under an optimized \texttt{vLLM} deployment, a single SOE intervention incurs only about 0.46s overhead, including approximately 0.21s for Teacher/Student sampling, 0.18s for embedding extraction, and 0.07s for scoring. Given an average end-to-end inference time of roughly 3.6s per problem in our setting, this corresponds to only a 12.8\% runtime overhead. Nevertheless, Table~\ref{tab:time_normalized} shows that SOE still achieves a higher density of correct solutions per unit wall-clock time than standard Self-Consistency across all five benchmarks.

Additional analyses, including orthogonality-score distributions, threshold-sensitivity results, and comprehensive ablations across components and model scales, are deferred to Appendix~\ref{app:exploration_efficiency_extra} and Appendix~\ref{sec:appendix_experiments}.

\subsection{Preliminary Results on Logic and Code Generation}
To further probe whether SOE extends beyond mathematical reasoning, we additionally evaluate it on ZebraLogic \cite{lin2025zebralogic} and HumanEvalPlus \cite{liu2023your}. As shown in Table~\ref{tab:domain_expansion}, SOE also improves problem-level accuracy on both tasks, suggesting that the benefit of orthogonal exploration is not limited to math benchmarks.
\begin{table}[h]
    \centering
    \small
    \begin{tabular}{lccc}
        \toprule
        \textbf{Benchmark} & \textbf{Baseline} & \textbf{Ours (SOE)} & \textbf{Rel. Imp.} \\
        \midrule
        ZebraLogic   & 56.23\% & \textbf{58.72\%} & +4.43\% \\
        HumanEvalPlus & 10.00\% & \textbf{16.67\%} & +66.7\% \\
        \bottomrule
    \end{tabular}
    \caption{Preliminary evaluation of SOE beyond mathematical reasoning.}
    \label{tab:domain_expansion}
\end{table}

\section{Analysis: Geometric Perspectives on Reasoning Collapse}

Based on our analysis, we view ``Reasoning Collapse'' as being associated with \textit{rank concentration} in the hidden-state covariance matrix.

In this section, we discuss two questions: (1) why LLM reasoning trajectories may become increasingly concentrated in low-dimensional subspaces, and (2) why our Orthogonal Latent Stitching (OLS) can act as a geometric intervention rather than simply adding new factual content.

\subsection{The Trap of Low-Rank Self-Consistency}

Let $H_t \in \mathbb{R}^{N \times d}$ be the matrix of hidden states in the context window at step $t$. We define the local covariance matrix $\Sigma_t = \frac{1}{N-1} H_t^T H_t$.
Reasoning collapse is associated with a decrease in the effective rank of $\Sigma_t$, i.e., $\mathrm{EffRank}(\Sigma_t) \ll d$, indicating that the states become increasingly concentrated in a subspace $\mathcal{S}$.

\textbf{Proposition 1 (Invariance of Subspace).} \textit{If the input context lies within a low-rank subspace $\mathcal{S}$, the output of a standard Self-Attention layer lies within the same subspace (or a linear transform thereof), limiting spontaneous exploration of its orthogonal complement.} \cite{dong2021attention,wang2020linformer}

The detailed proof and an analysis of why non-linearities (FFNs) may fail to substantially restore dimensionality are provided in Appendix \ref{app:proof_prop1}.

\subsection{Theoretical Justification for Progressive Rank Decay}

Why does the effective rank decrease over time during long chain-of-thought reasoning? We relate this phenomenon to a \textbf{spectrum contraction} effect of the Softmax operator in high-confidence regimes \cite{yang2017breaking}.

\textbf{Proposition 2 (Spectrum Contraction).} \textit{As the model becomes more confident (the norm of logits increases), the attention mechanism satisfies a non-expansion-style constraint on the dimensionality of the resulting representation, which can contribute to spectral concentration in the covariance matrix.}

Please refer to Appendix \ref{app:proof_thm2} for the formal proof. This result provides a theoretical perspective on why reasoning trajectories may become progressively concentrated over time, reducing exploration in directions useful for error correction.

\subsection{Orthogonal Injection: Geometry and Attention Rerouting}

A critical question arises: does SOE work because the Student provides \textit{new knowledge}, or because it introduces a geometric perturbation that changes the Teacher's subsequent trajectory? Our view is that the evidence more strongly supports the latter interpretation.

Consider the contribution to the spectrum in a direction $u_{\perp}$ lying in the orthogonal complement of the dominant subspace.
Prior to injection, the variance in this direction is $\sigma_{\perp}^2 \approx 0$.
After injecting an orthogonal vector $v_{\perp}$ (where $\|v_{\perp}\| > 0$), the covariance matrix acquires non-zero energy in this direction, yielding

\begin{equation}
    \mathrm{rank}(\Sigma_{\text{new}}) \ge \mathrm{rank}(\Sigma_{\text{old}})
\end{equation}

and, in the idealized full-rank-augmentation case,

\begin{equation}
    \mathrm{rank}(\Sigma_{\text{new}}) = \mathrm{rank}(\Sigma_{\text{old}}) + 1.
\end{equation}

This change does not by itself guarantee better reasoning, but it provides a geometric mechanism by which the trajectory can access previously underexplored directions. In practice, we hypothesize that semantically structured Student probes are helpful not merely because they add tokens, but because they introduce heterogeneous signals that reroute the Teacher's attention away from its current biased trajectory. From this perspective, SOE can be understood as a geometric intervention that expands the space of subsequent exploration, rather than purely as a knowledge-injection mechanism.

\section{Conclusion}
We presented evidence that \textit{Reasoning Collapse} is associated with geometric concentration into a low-rank \textit{Bias Manifold}, which can reduce exploration toward corrective directions in the orthogonal complement. To address this, we introduced \textbf{Spectral Orthogonal Exploration (SOE)}, a ``Student Guides Teacher'' framework that uses \textit{Orthogonal Latent Stitching} to intervene on biased reasoning trajectories. Our empirical and theoretical analyses suggest that SOE is effective partly because it expands the geometric space available for subsequent exploration, rather than solely because it injects additional information. More broadly, these findings suggest that mitigating reasoning bottlenecks may benefit from going beyond probabilistic sampling and incorporating targeted geometric interventions in latent space.

\section{Limitations}

While our proposed Spectral Orthogonal Exploration (SOE) framework demonstrates significant efficacy in mitigating reasoning collapse, we acknowledge several limitations that merit further investigation.

First, although SOE improves sample efficiency by requiring fewer total tokens to discover correct solutions, it still introduces additional wall-clock latency due to the Monte Carlo look-ahead, hidden-state extraction, and scoring steps required to estimate the local bias manifold and select orthogonal probes.

Second, the core mechanism of SOE relies on explicitly accessing and manipulating the model's internal hidden states to compute effective rank and orthogonal residuals. This requirement restricts the applicability of our method to open-weights models, precluding its direct use with closed-source LLMs where only API access is available.

Finally, although we include additional validation on larger Teacher models and preliminary experiments beyond mathematics, our empirical study remains centered on mathematical reasoning, and a full-scale evaluation across broader domains and model families is still needed.



\bibliography{custom}

\appendix

\section{Detailed Example of Spectral Orthogonal Exploration}
\label{app:soe_example}

In this section, we provide a step-by-step walk-through of the SOE pipeline on a specific problem from the AIME 2025 dataset. We visualize the breakdown of the Teacher's reasoning collapse and how the Student's orthogonal probe successfully ejects the generation from the Bias Manifold.

\subsection{Problem Input}
\begin{tcolorbox}[colback=gray!10, colframe=gray!50, title=\textbf{Input Prompt}]
\textbf{System Prompt:} Please reason step by step, and put your final answer within \verb|\boxed{}|.

\textbf{User Problem:} 
In $\\triangle ABC$ points $D$ and $E$ lie on $\\overline{AB}$ so that $AD < AE < AB$, while points $F$ and $G$ lie on $\\overline{AC}$ so that $AF < AG < AC$. Suppose $AD = 4$, $DE = 16$, $EB = 8$, $AF = 13$, $FG = 52$, and $GC = 26$. Let $M$ be the reflection of $D$ through $F$, and let $N$ be the reflection of $G$ through $E$. The area of quadrilateral $DEGF$ is $288$. Find the area of heptagon $AFNBCEM$.

\end{tcolorbox}

\subsection{Baseline: Reasoning Collapse (Greedy Decoding)}
The Teacher model ($\pi_\theta$) initially fails to solve the problem. Following is the generation.

We are given:

\begin{itemize}
    \item Points $D, E$ on $AB$ such that $AD < AE < AB$,
    \item Points $F, G$ on $AC$ such that $AF < AG < AC$,
    \item Lengths:
    \begin{itemize}
        \item $AD = 4$
        \item $DE = 16$
        \item $EB = 8$
        \item $AF = 13$
        \item $FG = 52$
        \item $GC = 26$
    \end{itemize}
    \item $M$ is the reflection of $D$ through $F$,
    \item $N$ is the reflection of $G$ through $E$,
    \item The area of quadrilateral $DEGF$ is $288$,
    \item Find the area of the heptagon $AFNBCEM$.
\end{itemize}

\bigskip
\textbf{Step 1: Understand the given structure}

\medskip
\textbf{On segment $AB$:}
\[
AD = 4, \quad DE = 16, \quad EB = 8
\]
\[
AB = AD + DE + EB = 4 + 16 + 8 = 30
\]

\medskip
\textbf{On segment $AC$:}
\[
AF = 13, \quad FG = 52, \quad GC = 26
\]
\[
AC = AF + FG + GC = 13 + 52 + 26 = 91
\]

\bigskip
\textbf{Step 2: Understanding points $M$ and $N$}

\begin{itemize}
    \item $M$ is the reflection of $D$ through point $F$, meaning $F$ is the midpoint of segment $DM$.
    \item $N$ is the reflection of $G$ through point $E$, meaning $E$ is the midpoint of segment $GN$.
\end{itemize}

Since $D$ lies on $AB$ while $F$ lies on $AC$, the reflection is a point reflection in the plane, not along a line.
Similarly, $N$ need not lie on any side of triangle $ABC$.

The vertices of the required heptagon are
\[
A, F, N, B, C, E, M.
\]

\bigskip
\textbf{Step 3: Strategy}

We aim to express the area of the heptagon $AFNBCEM$ in terms of the area of triangle $ABC$.
However, since points $M$ and $N$ lie outside triangle $ABC$, the heptagon extends beyond the triangle, and we cannot simply subtract interior regions.

Thus, we adopt a vector-area approach.

\bigskip
\textbf{Step 4: Vector setup}

Let
\[
\vec{AB} = \vec{b}, 
\qquad
\vec{AC} = \vec{c}.
\]

Then:
\[
| \vec{b} \times \vec{c} | = AB \cdot AC \cdot \sin \theta = 30 \cdot 91 \cdot \sin \theta = 2730 \sin \theta.
\]

Points along $AB$:
\[
\vec{D} = \frac{4}{30} \vec{b} = \frac{2}{15} \vec{b},
\qquad
\vec{E} = \frac{20}{30} \vec{b} = \frac{2}{3} \vec{b}.
\]

Points along $AC$:
\[
\vec{F} = \frac{13}{91} \vec{c} = \frac{1}{7} \vec{c},
\qquad
\vec{G} = \frac{65}{91} \vec{c} = \frac{5}{7} \vec{c}.
\]

\bigskip
\textbf{Step 5: Area of quadrilateral $DEGF$}

Using the polygon area formula,
\[
[\!DEGF] 
= \frac12 
\left|
\vec{D} \times \vec{E}
+ \vec{E} \times \vec{G}
+ \vec{G} \times \vec{F}
+ \vec{F} \times \vec{D}
\right|.
\]

Compute each term:
\[
\vec{D} \times \vec{E} = 0,
\quad
\vec{G} \times \vec{F} = 0,
\]
\[
\vec{E} \times \vec{G} = \frac{10}{21} (\vec{b} \times \vec{c}),
\quad
\vec{F} \times \vec{D} = -\frac{2}{105} (\vec{b} \times \vec{c}).
\]

Thus,
\[
[\!DEGF] 
= \frac12 \cdot \frac{16}{35} |\vec{b} \times \vec{c}|
= \frac{8}{35} \cdot 2730 \sin \theta
= 624 \sin \theta.
\]

Given
\[
[\!DEGF] = 288,
\]
we obtain
\[
\sin \theta = \frac{288}{624} = \frac{6}{13}.
\]

\bigskip
\textbf{Step 6: Area of triangle $ABC$}

\[
[\triangle ABC]
= \frac12 \cdot 30 \cdot 91 \cdot \sin \theta
= 1365 \cdot \frac{6}{13}
= 630.
\]

\bigskip
\textbf{Conclusion}

The remaining task is to express the area of heptagon $AFNBCEM$ in terms of the area of triangle $ABC$, using reflections about points $F$ and $E$.

\subsection{Phase 1: Truncation Points}
We detect the potential collapse and truncate the erroneous trajectory at three distinct milestones ($t_1, t_2, t_3$).

\begin{itemize}

\item \textbf{Truncation Point 1 ($t_1$):}

We are given:

\begin{itemize}
    \item Points $D, E$ on $AB$ such that $AD < AE < AB$,
    \item Points $F, G$ on $AC$ such that $AF < AG < AC$,
    \item Lengths:
    \begin{itemize}
        \item $AD = 4$
        \item $DE = 16$
        \item $EB = 8$
        \item $AF = 13$
        \item $FG = 52$
        \item $GC = 26$
    \end{itemize}
    \item $M$ is the reflection of $D$ through $F$,
    \item $N$ is the reflection of $G$ through $E$,
    \item Area of quadrilateral $DEGF = 288$,
    \item We are to find the area of heptagon $AFNBCEM$.
\end{itemize}

\bigskip
\textbf{Step 1:}

\hrulefill

\item \textbf{Truncation Point 2 ($t_2$):}

We are given:

\begin{itemize}
    \item Points $D, E$ on $AB$ such that $AD < AE < AB$,
    \item Points $F, G$ on $AC$ such that $AF < AG < AC$,
    \item Lengths:
    \begin{itemize}
        \item $AD = 4$
        \item $DE = 16$
        \item $EB = 8$
        \item $AF = 13$
        \item $FG = 52$
        \item $GC = 26$
    \end{itemize}
    \item $M$ is the reflection of $D$ through $F$,
    \item $N$ is the reflection of $G$ through $E$,
    \item Area of quadrilateral $DEGF = 288$,
    \item We are to find the area of heptagon $AFNBCEM$.
\end{itemize}

\bigskip
\textbf{Step 1: Understand the given structure}

\medskip
\textbf{On $AB$:}
\[
AD = 4,\quad DE = 16,\quad EB = 8
\]
\[
AB = AD + DE + EB = 4 + 16 + 8 = 30
\]

\medskip
\textbf{On $AC$:}
\[
AF = 13,\quad FG = 52,\quad GC = 26
\]
\[
AC = AF + FG + GC = 13 + 52 + 26 = 91
\]

\bigskip
\textbf{Step 2:}

\hrulefill

\item \textbf{Truncation Point 3 ($t_3$):}

We are given:

\begin{itemize}
    \item Points $D, E$ on $AB$ such that $AD < AE < AB$,
    \item Points $F, G$ on $AC$ such that $AF < AG < AC$,
    \item Lengths:
    \begin{itemize}
        \item $AD = 4$
        \item $DE = 16$
        \item $EB = 8$
        \item $AF = 13$
        \item $FG = 52$
        \item $GC = 26$
    \end{itemize}
    \item $M$ is the reflection of $D$ through $F$,
    \item $N$ is the reflection of $G$ through $E$,
    \item Area of quadrilateral $DEGF = 288$,
    \item We are to find the area of heptagon $AFNBCEM$.
\end{itemize}

\bigskip
\textbf{Step 1: Understand the given structure}

\medskip
\textbf{On $AB$:}
\[
AB = 4 + 16 + 8 = 30
\]

\medskip
\textbf{On $AC$:}
\[
AC = 13 + 52 + 26 = 91
\]

\bigskip
\textbf{Step 2: Understand points $M$ and $N$}

\begin{itemize}
    \item $M$ is the reflection of $D$ through $F$, so $F$ is the midpoint of $DM$.
    \item $N$ is the reflection of $G$ through $E$, so $E$ is the midpoint of $GN$.
\end{itemize}

Since $D$ lies on $AB$ and $F$ lies on $AC$, the reflection is a point reflection in the plane, not along a line.

We are asked to compute the area of the heptagon:
\[
AFNBCEM
\]

\bigskip
\textbf{Step 3:}

\end{itemize}

\subsection{Phase 2: Micro-SVD \& Probe Selection (At Truncation Point $t_1$)}
Here we detail the geometric intervention at the first truncation point.

\subsubsection{Manifold Estimation (Monte Carlo Samples)}
Taking truncation point 1 as an example, the Teacher generates $N$ look-ahead paths to estimate the local covariance matrix $\Sigma_{t_1}$. Note how these samples are semantically similar (confirming the Bias Manifold).

\begin{quote}
\textit{MC Sample 1:} Understand the configuration.  
We have triangle \\

\textit{MC Sample 2:} Understand the configuration.  
We have triangle  \\

\textit{MC Sample 3:} Understand the configuration.  
We have triangle  \\

\textit{MC Sample 4:} Understand and use given lengths.  
We  \\

\textit{MC Sample 5:} Understand the configuration.  
We are dealing with  \\

\textit{MC Sample 6:} Understand the configuration.  
We have triangle  \\

\textit{MC Sample 7:} Understand the configuration.  
We have triangle  \\

\textit{MC Sample 8:} Understand the configuration.  
Let’s first sketch 
\end{quote}

\subsubsection{Student Probe Candidates ($L=8$ tokens)}
The Student model ($\pi_\phi$) generates 8 candidate sequences. We calculate the \textbf{Orthogonality Score} for each relative to the Teacher's estimated manifold.
\begin{quote}
\textit{MC Sample 1:} Find $AE, AB, AC$.  
\hfill Orthogonality Score: 0.9403

\textit{MC Sample 2:} Let $A(x_1$.  
\hfill Orthogonality Score: 0.9844

\textit{MC Sample 3:} Let $A = (0$.  
\hfill Orthogonality Score: 0.9802

\textit{MC Sample 4:} First, let's find the  
\hfill Orthogonality Score: 0.7196

\textit{MC Sample 5:} Let $A = (0,$.  
\hfill Orthogonality Score: 0.9801

\textit{MC Sample 6:} We are given the lengths $AD$.  
\hfill Orthogonality Score: 0.8500

\textit{MC Sample 7:} We know $AB = AD + $.  
\hfill Orthogonality Score: 0.9643

\textbf{\textit{MC Sample 8:}} \textbf{Since $AD = 4$, \dots}  
\hfill \textbf{Orthogonality Score: 0.9724}
\end{quote}


\section{Reasoning Collapse Evidence}

\begin{figure}[h]
    \centering
    \includegraphics[width=0.75\linewidth]{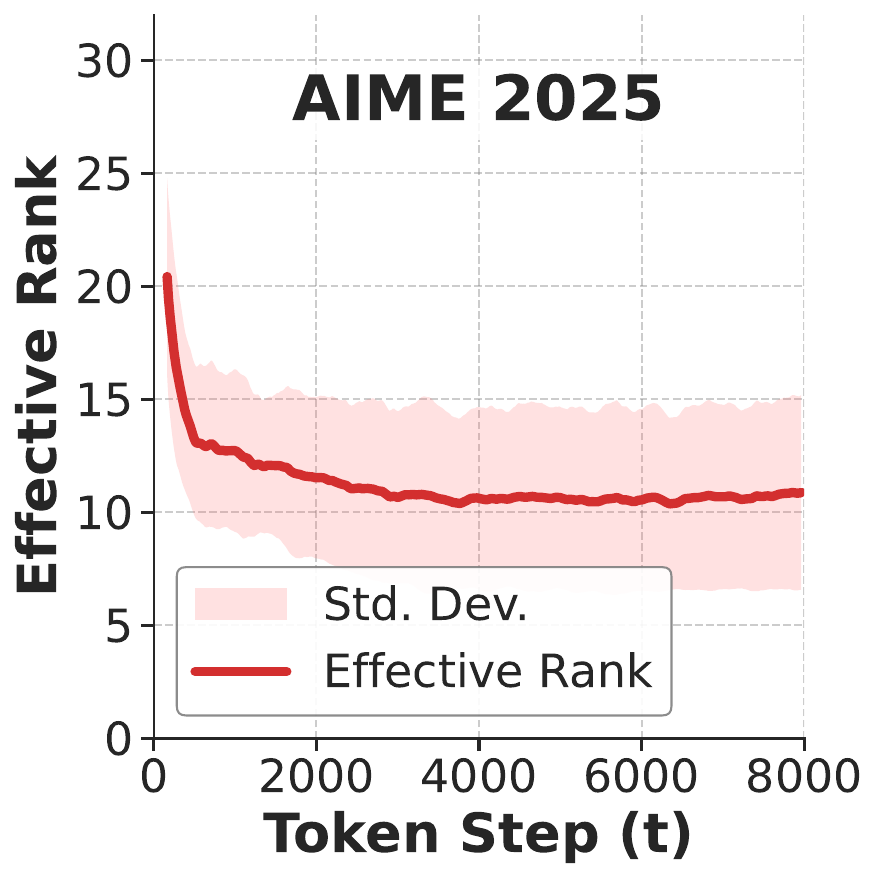}
    \caption{Reasoning Collapse of 8B Model}

        \label{fig:8B}
\end{figure}

\begin{figure}[h]
        \centering
        \includegraphics[width=0.75\linewidth]{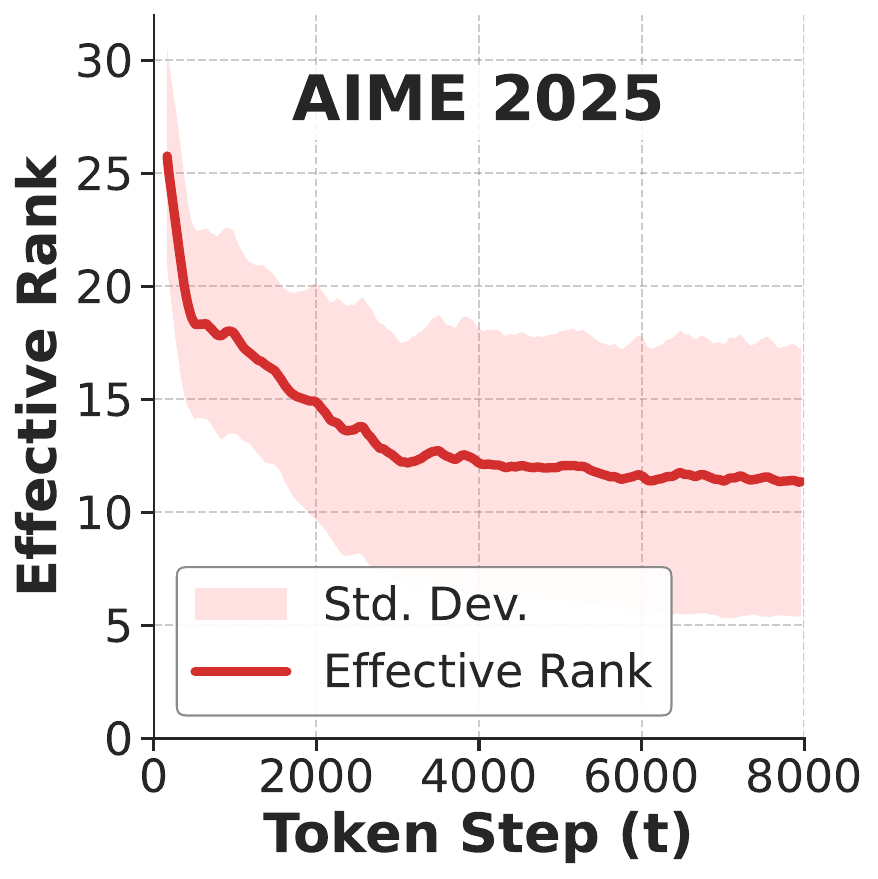}
        \caption{Reasoning Collapse of 32B Model}
        \label{fig:32B}
\end{figure}

To verify the universality of the observed geometric pathology, we extended our spectral analysis to models of varying scales, specifically \texttt{Qwen3-8B}\cite{qwen3technicalreport} and \texttt{QwQ-32B}\cite{qwq32b}. Employing the identical methodology on the AIME 2025 benchmark, we tracked the effective rank dynamics of the hidden state covariance matrices during inference. As illustrated in Figure~\ref{fig:8B} and Figure~\ref{fig:32B}, both models exhibit a consistent pattern of spectral degeneration: despite their increased parameter counts, the effective rank decays significantly as the reasoning chain lengthens and enters erroneous loops. This evidence suggests that reasoning collapse is a fundamental geometric phenomenon that persists across model scales, rather than an artifact specific to smaller architectures.
\label{sec:Reasoning Collapse Evidence}
\label{sec:Collapse}

\section{Ablation Studies and Robustness Analysis}
\label{sec:appendix_experiments}

\subsection{Ablation Studies}
\label{app:ablation}

To rigorously evaluate the contribution of each component in the Spectral Orthogonal Exploration (SOE) framework, we conduct ablation studies focusing on the necessity of the probe injection and the effectiveness of the geometric selection mechanism. Due to computational resource constraints, these analyses were performed on the AIME 2025 benchmark.

\paragraph{Validity of the Baseline as an Ablation.}
We first emphasize that the \textbf{Baseline (Self-Consistency)} reported in the main text serves as the fundamental ablation of our framework. It represents the "No Injection" setting, where the Teacher model samples repeatedly from the same collapsed trajectory without any external intervention. The significant performance gap between the Baseline and SOE (as shown in the main results) confirms the essential value of the "Student Guides Teacher" paradigm.

\paragraph{Effectiveness of Micro-SVD Selection (Random vs. Orthogonal).}
To isolate the impact of our \textit{Orthogonal Latent Stitching} (Micro-SVD) mechanism, we conducted a further ablation. In this setting, the Student generates candidate probes, but instead of selecting the candidate with the highest orthogonality score, we select one \textbf{at random} to stitch into the Teacher's context.

This comparison allows us to determine whether performance gains arise merely from adding "noise" (random perturbation) or from the specific geometric directionality of our method. As shown in Table~\ref{tab:ablation}, while random injection provides a performance boost over the baseline (58.82\% vs 35.29\%) by disrupting the bias manifold, our proposed SOE method achieves a significantly higher accuracy (70.59\%). This result confirms that the Micro-SVD mechanism effectively identifies high-value directions in the orthogonal complement that are superior to random perturbations.

\begin{table}[h]
    \centering
    \small
    \begin{tabular}{l c c}
    \toprule
    \textbf{Method}  & \textbf{Pass@16 (AIME 25)} \\
    \midrule
    Baseline  & 35.29\% \\
    Ablation  & 58.82\% \\
    \rowcolor{gray!10} \textbf{Ours (SOE)}  & \textbf{70.59\%} \\
    \bottomrule
    \end{tabular}
    \caption{Ablation study on AIME 2025. "Random Probe" indicates selecting a Student trace without geometric filtering. The results demonstrate that while any external diversity helps, directed orthogonal exploration yields the optimal performance.}
    \label{tab:ablation}
\end{table}

\begin{figure}[h]
    \centering
    \includegraphics[width=0.8\linewidth]{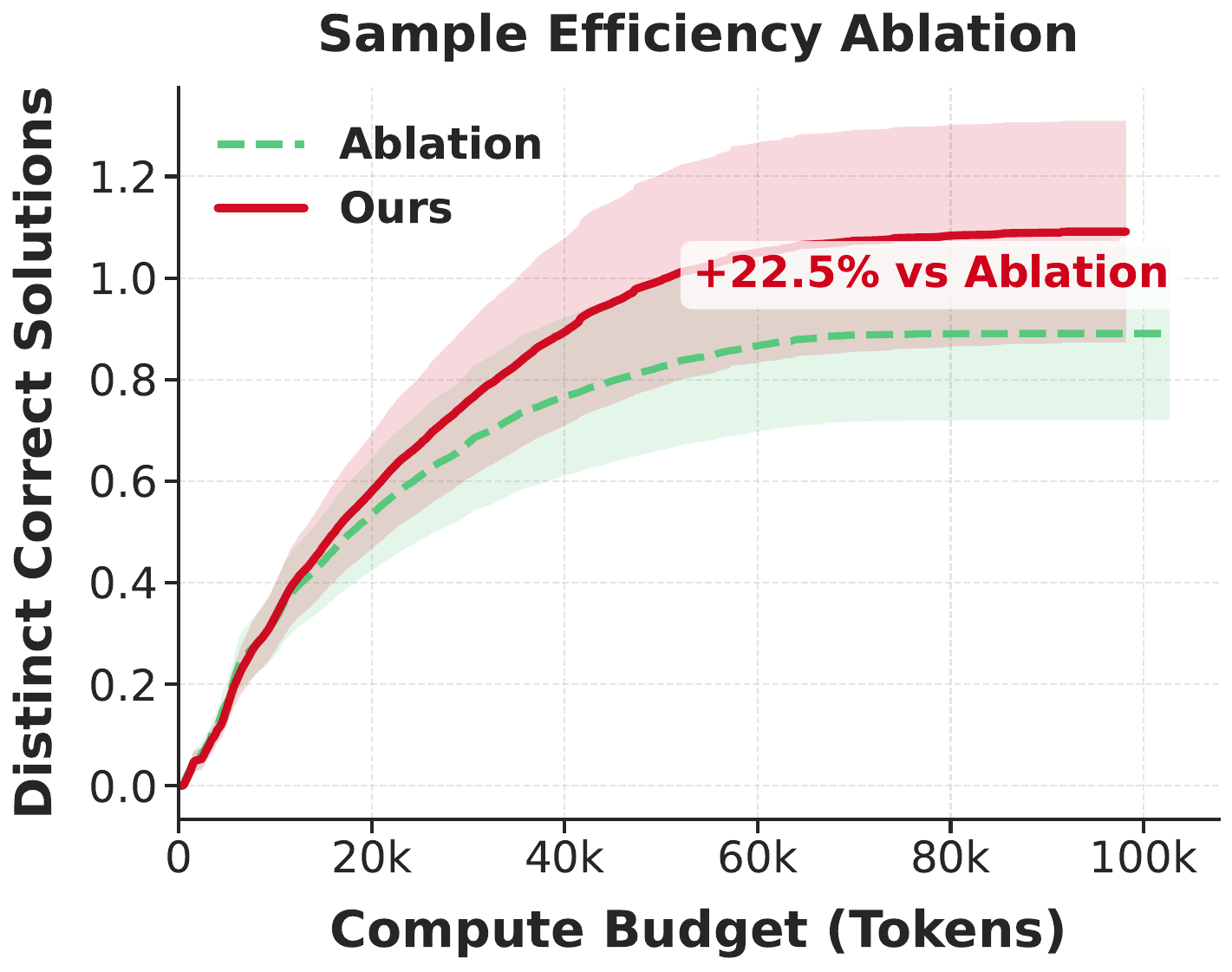} 
    \caption{Sampling efficiency comparison between Baseline, Random Ablation, and SOE on AIME 2025. SOE discovers correct solutions at a significantly faster rate.}
    \label{fig:ablation_curve}
\end{figure}

\paragraph{Sampling Efficiency.}
Figure~\ref{fig:ablation_curve} further illustrates the sampling efficiency comparison. Our method reaches high-confidence solutions with fewer samples compared to both the Baseline and the Random Ablation, indicating that SOE navigates the solution space more efficiently.

\subsection{Additional Analysis of Exploration Efficiency}
\label{app:exploration_efficiency_extra}

\paragraph{Orthogonality Score Distribution.}
To validate the geometric mechanism underlying SOE, we analyze the distribution of Orthogonality Scores for Student-generated probes across benchmarks.

\begin{figure}[t]
    \centering
    \includegraphics[width=0.75\linewidth]{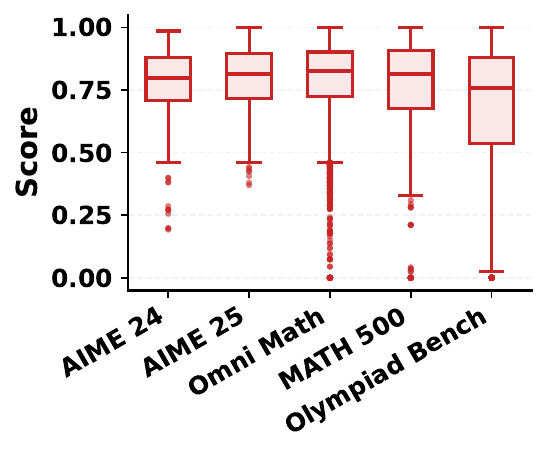}
    \caption{\textbf{Distribution of Orthogonality Scores} for Student-generated probes across benchmarks. The score ranges from $0$ to $1$, where a value of $1$ indicates perfect orthogonality relative to the Teacher's Bias Manifold, and $0$ denotes complete alignment (collinearity).}
    \label{fig:our_score}
\end{figure}

As shown in Figure~\ref{fig:our_score}, the Student-generated probes consistently exhibit high orthogonality relative to the Teacher's local Bias Manifold. This supports our claim that the Student provides sufficiently heterogeneous reasoning signals to project into the Teacher's orthogonal complement, thereby helping escape low-rank confinement.

\subsection{Sensitivity Analysis of the Diversity Threshold}
\label{app:threshold_sensitivity}

To test the robustness of the \textit{Distinct Correct Solutions} metric, we vary the cosine-similarity threshold $\tau$ used to distinguish semantically different correct solutions. Table~\ref{tab:threshold_sensitivity} reports the results on AIME 2024 and AIME 2025. Across all threshold choices, SOE consistently discovers more distinct correct solutions than the baseline, confirming that the improvement in exploration efficiency is robust to the specific choice of $\tau$.

\begin{table*}[t]
    \centering
    \small
    \setlength{\tabcolsep}{6pt}
    \begin{tabular}{lcccc}
        \toprule
        \textbf{Cosine Threshold} & \textbf{AIME 25 (Baseline)} & \textbf{AIME 25 (Ours)} & \textbf{AIME 24 (Baseline)} & \textbf{AIME 24 (Ours)} \\
        \midrule
        0.80 (Strict)      & 0.244 & \textbf{0.886} & 0.265 & \textbf{0.944} \\
        0.90               & 0.356 & \textbf{1.660} & 0.495 & \textbf{1.971} \\
        0.95 (Original)    & 0.640 & \textbf{2.482} & 1.028 & \textbf{3.258} \\
        1.00 (Exact Match) & 0.740 & \textbf{3.080} & 1.028 & \textbf{4.285} \\
        \bottomrule
    \end{tabular}
    \caption{Sensitivity analysis of the diversity threshold $\tau$ used in the \textit{Distinct Correct Solutions} metric. SOE consistently finds more distinct correct solutions than the baseline across all tested thresholds.}
    \label{tab:threshold_sensitivity}
\end{table*}

\subsection{Robustness Analysis}

To evaluate the stability of our method against stochastic variations, we conducted repeated experiments on the AIME 2025 benchmark using three distinct random seeds ($\{42, 1024, 2024\}$). Due to computational resource constraints, this robustness analysis was restricted to the AIME 2025 dataset. As shown in Table~\ref{tab:robustness}, our method (SOE) consistently outperforms the baseline across all random seeds, demonstrating that the performance improvements are not artifacts of a specific initialization. Notably, while the baseline fluctuates between $29.41\%$ and $41.18\%$, our method achieves a stable accuracy of up to $70.59\%$.

\begin{table}[h]
    \centering
    \small
    \begin{tabular}{c c c c}
    \toprule
    \textbf{Seed} & \textbf{Baseline} & \textbf{Ours (SOE)}  \\
    \midrule
    42 & 29.41\% & 64.71\%  \\
    1024 & 41.18\% & 70.59\% \\
    2024 & 35.29\% & 70.59\%  \\
    \bottomrule
    \end{tabular}
    \caption{Robustness analysis on AIME 2025 across three different random seeds. Our method demonstrates consistent performance gains despite stochastic variations.}
    \label{tab:robustness}
\end{table}

\section{Theoretical Proofs and Extended Analysis}

\subsection{Proof of Proposition 1 and Analysis of Non-Linearity}
\label{app:proof_prop1}

\begin{proof}
Consider a simplified linear attention head. The output for a query $q$ is given by:
\begin{equation}
    \mathrm{Attn}(q, K, V) = \sum_{i=1}^N \mathrm{softmax}(q^T k_i) v_i
\end{equation}
where $v_i = W_V h_i$. Since each input $h_i \in \mathcal{S}$, each value vector $v_i$ lies in $W_V \mathcal{S}$. The output is therefore a convex combination of vectors in $W_V \mathcal{S}$. Thus, the generated token embedding remains within the span of the existing history (up to the corresponding linear transform). Without an external perturbation or a mechanism that introduces new directions, the model has limited ability to explore the orthogonal complement $\mathcal{S}^\perp$.
\end{proof}

\paragraph{Remark 1: The Limited Role of Pointwise Non-Linearity.}
A natural counter-argument is that the Feed-Forward Networks (FFNs) in Transformers, which contain non-linear activation functions (such as GELU or ReLU), might restore lost dimensionality. We analyze this possibility by examining the coordinate-wise nature of these activations.

Consider the FFN update: $h' = W_2 \sigma(W_1 h)$. The linear transformations $W_1, W_2$ satisfy $\mathrm{rank}(Wh) \leq \mathrm{rank}(h)$, so any dimensional expansion would need to arise from the non-linearity $\sigma(\cdot)$.

However, standard activation functions are \textbf{coordinate-wise}: $\sigma(v)_i = \sigma(v_i)$. They operate on each dimension independently without mixing information.

\begin{itemize}
    \item \textbf{Zero-Variance Persistence}: If the input distribution has collapsed such that variance is zero along a basis vector $e_j$, then for any sample $h$, the $j$-th component is constant (or zero). Since $\sigma$ acts independently, the output $\sigma(h_j)$ remains constant. The activation does not by itself introduce variation in that direction.
    \item \textbf{Manifold Curvature vs. Expansion}: While non-linearity can bend a flat $k$-dimensional subspace into a curved $k$-dimensional manifold, it does not necessarily increase the \textit{intrinsic dimensionality}. A piece of paper crumpled into a ball still has intrinsic dimension 2; it does not acquire volume.
\end{itemize}

Consequently, without an external geometric intervention, the FFN may warp the collapsed Bias Manifold but does not necessarily restore exploration in previously inactive directions \cite{geva2021transformer}.

\subsection{Proof of Proposition 2}
\label{app:proof_thm2}

\begin{proof}
Let the attention weights be $A \in \mathbb{R}^{N \times N}$, and consider the simplified update rule
\begin{equation}
    H_{t+1} = A H_t.
\end{equation}

In later stages of reasoning, the model may exhibit a form of confirmation bias in which many tokens attend heavily to a small set of anchor tokens. In this regime, the rows of the attention matrix can become increasingly similar, reducing the diversity of directions that contribute to the next hidden-state update.

Using the singular-value inequality $\sigma_i(XY) \le \|X\| \sigma_i(Y)$, we obtain that
\begin{equation}
    \mathrm{rank}(H_{t+1}) \le \min\big(\mathrm{rank}(A), \mathrm{rank}(H_t)\big).
\end{equation}
Therefore, the attention update satisfies a non-expansion-style constraint on algebraic rank. When combined with increasingly concentrated attention patterns, this provides a theoretical perspective on why the representation may become progressively more spectrally concentrated over time.
\end{proof}

\subsection{Semantic Sparsity Analysis}
\label{app:sparsity}

Here we discuss the informational contribution of the injection. Let the injected Student sequence be $s_{\mathrm{inj}}$ with length $L_{\mathrm{inj}}$ (8 tokens in our experiments), and let the Teacher's context length be $L_{\mathrm{ctx}}$ (8192 tokens in our experiments).

\textbf{Argument: Semantic Sparsity.}
The direct contribution of the injection to the global context mean is small:
\begin{equation}
    \Delta \mu \approx \frac{L_{\mathrm{inj}}}{L_{\mathrm{ctx}}} (h_{\mathrm{Student}} - \mu_{\mathrm{Teacher}}) \approx 0
\end{equation}
Since $L_{\mathrm{inj}} \ll L_{\mathrm{ctx}}$, the injected sequence contributes only a limited shift to the global centroid of the context. This observation is consistent with the view that SOE's gains are not solely explained by large-scale information transfer through the added tokens. Instead, the injected probe may be more effective as a localized geometric perturbation that changes subsequent attention routing and exploration.

\section{LLM Usage Statement}
\label{LLM Usage Statement}
Large Language Models (LLMs) such as ChatGPT and Gemini were used as general-purpose writing assistants to improve readability and clarity of the manuscript (\emph{e.g.}, grammar checking, LaTeX formatting, and sentence restructuring).
LLMs were not used to determine the research idea, perform literature search or generate the paper's key insights and conclusions.
All analyses and conclusions are solely those of the authors.

\end{document}